\documentclass[conference]{IEEEtran}
\IEEEoverridecommandlockouts
\usepackage{cite}
\usepackage{amsmath,amssymb,amsfonts}
\usepackage{algorithmic}
\usepackage{graphicx}
\usepackage{textcomp}
\usepackage{xcolor}
\usepackage{booktabs}

\def\BibTeX{{\rm B\kern-.05em{\sc i\kern-.025em b}\kern-.08em
    T\kern-.1667em\lower.7ex\hbox{E}\kern-.125emX}}

\begin{document}

\title{Interpretability-Guided Auditing for Clinical Language Models for Discharge Prediction}

\author{
\IEEEauthorblockN{
Ha Na Cho\IEEEauthorrefmark{1},
Sairam Sutari\IEEEauthorrefmark{1},
Alexander Lopez\IEEEauthorrefmark{2},
Hansen Bow\IEEEauthorrefmark{2},
Kai Zheng\IEEEauthorrefmark{1}
}
\IEEEauthorblockA{\IEEEauthorrefmark{1}
Informatics, Computer Science\\
University of California, Irvine\\
Irvine, CA, USA
}
\IEEEauthorblockA{\IEEEauthorrefmark{2}
Department of Neurosurgery\\
University of California, Irvine\\
Irvine, CA, USA
}
}

\maketitle

\begin{abstract}
Clinical natural language processing (NLP) models for hospital discharge prediction are vulnerable to explicit discharge-related lexical cues in clinical notes, even under temporal filtering of note timestamps. These cues can act as proxies for future outcomes and lead models to rely on shortcut-like signals. We propose AuditBERT-Discharge, a lightweight framework that reframes interpretability as a training-time auditing mechanism. The token-level SHAP from an initial BioClinicalBERT model is used to identify highly attributed discharge-related tokens, which are then masked in the training corpus prior to retraining. We evaluate the approach on a retrospective cohort of elective spine surgery encounters from a single academic medical center. Qualitative attribution analysis and preliminary evaluation suggest that SHAP-guided auditing reduces reliance on explicit discharge-proxy language and produces more conservative and better-calibrated predictions, although not uniformly higher discrimination performance. These findings position interpretability-guided auditing as a practical strategy for identifying and constraining shortcut-like lexical dependencies in temporally sensitive clinical NLP tasks.
\end{abstract}


\section{Introduction}

Predicting near-term hospital discharge from electronic health records (EHRs) is a clinically valuable yet methodologically challenging task. In particular, models trained on clinical notes may become susceptible to implicit temporal leakage, where discharge-related phrases appear in notes written close to or after the decision point, potentially inflating predictive performance. While recent clinical NLP studies report strong performance using large language models or fine-tuned transformers, such gains often obscure whether models rely on clinically valid signals or spurious lexical shortcuts \cite{b1,b2}.

Existing approaches \cite{b3,b4} typically treat interpretability as a post-hoc explanatory tool, rather than as a mechanism to enforce input validity. As a result, explanation methods such as Shapley Additive exPlanations (SHAP) are often used to rationalize predictions after training, without preventing models from learning leakage-prone cues in the first place. This raises broader concerns about robustness, generalizability, and deployability in real clinical settings. For example, progress notes authored on the operative day frequently contain phrases such as ``anticipate discharge tomorrow,'' ``likely home POD1,'' or abbreviated tokens such as ``dc'' or ``DC.'' Although such notes precede the formal discharge event, they encode downstream clinical decisions and act as direct proxies for the target label. Models trained on these notes can therefore achieve inflated performance without learning clinically meaningful predictors of readiness for discharge.

We study this problem in a retrospective cohort of elective spine surgery encounters from a large academic medical center, where next-day discharge prediction is performed using operative-day clinical notes under strict temporal filtering. In this paper, we investigate whether interpretability can be used not only for post-hoc explanation but also as a mechanism for auditing model inputs. We introduce AuditBERT-Discharge, a lightweight framework in which token-level SHAP attribution from an initial BioClinicalBERT model is used to identify highly attributed discharge-related lexical cues. These cues are then masked in the training corpus prior to retraining an otherwise identical model. Our goal is not to maximize predictive performance, but to examine whether interpretability-guided auditing can reduce reliance on explicit discharge-proxy language cues and produce more conservative prediction behavior in temporally sensitive clinical NLP tasks.

Our contribution is intentionally scoped: we demonstrate that SHAP-guided lexical auditing reveals patterns of over-reliance on discharge cues in clinical notes and that retraining on an audited input space yields models that exhibit more conservative and better-calibrated prediction behavior. This work is not a full-scale modeling benchmark, but a focused methodological demonstration illustrating how interpretability can be operationalized as a pre-hoc control mechanism in clinical NLP. Rather than positioning AuditBERT-Discharge as a competitive benchmark against alternative architectures, this work focuses on auditing patterns within a fixed model class to isolate the effect of interpretability-guided suppression of temporal leakage. We evaluate this approach using retrospective EHR data from 1,956 elective spine surgery encounters at a large academic medical center.

\section{Related Work}

\subsection{Discharge Prediction with Clinical Text}

Early work on discharge and length-of-stay prediction relied primarily on structured EHR data, including demographics, diagnoses, procedures, and laboratory results, using classical machine learning models \cite{b5,b6}. These approaches demonstrated that structured variables alone can achieve reasonable discrimination for discharge-related outcomes, particularly in surgical populations. More recent studies have incorporated unstructured clinical notes using neural- and transformer-based models such as ClinicalBERT and BioClinicalBERT, often improving predictive performance \cite{b7}. However, note-based models are vulnerable to temporal leakage, as clinical documentation can implicitly encode discharge decisions or future events. Multimodal approaches combining structured and unstructured data further improve performance, but most prior work emphasizes accuracy and provides limited evaluation of temporal validity or clinical appropriateness of learned signals \cite{b8}.

\subsection{Interpretability in Clinical Prediction Models}

Interpretability has become a central concern in clinical machine learning due to the high-stakes nature of medical decision-making. Post-hoc explanation methods such as SHAP, LIME, attention visualization, and gradient-based saliency maps are widely used to identify influential features or tokens in trained models \cite{b3,b9}. In clinical NLP, these methods are typically applied to explain individual predictions or to provide retrospective information on model behavior \cite{b4}.

Beyond explanation, several studies have used interpretability for qualitative inspection, bias analysis, and dataset auditing, revealing spurious correlations, annotation artifacts, or shortcut learning in both general-domain and clinical models \cite{b10,b11,b12}. In healthcare settings, interpretability analyses have increasingly been used to inspect potential biases and spurious correlations in machine learning models, though they are still primarily applied as retrospective diagnostic tools rather than training-time control mechanisms.

Despite their widespread use, interpretability methods are most often applied after model training and evaluation, where they primarily serve a descriptive role. While some studies use explanations for qualitative inspection, bias analysis, or model debugging, interpretability is rarely integrated into the modeling pipeline as a mechanism for enforcing temporal validity or preventing the learning of spurious or leakage-prone signals, particularly in clinical text \cite{b10}.

\subsection{Temporal Leakage and the Need for Auditing}

Temporal leakage is a well-recognized challenge in clinical prediction tasks, particularly those involving future outcomes such as discharge timing. Leakage can arise when models inadvertently access information recorded after the prediction window or when proxy variables encode downstream clinical decisions. In note-based models, this issue is especially acute, as discharge-related language and planning cues may appear in progress notes before formal discharge.

Most prior work mitigates temporal leakage through rule-based preprocessing, such as timestamp filtering, excluding discharge summaries, or removing explicit discharge terms \cite{b13}. While effective against overt leakage, these heuristics do not systematically detect indirect proxy signals or documentation artifacts that may still influence model predictions. To our knowledge, no prior discharge prediction work operationalizes interpretability as a pre-training auditing mechanism to govern model inputs. We address this gap by reframing interpretability from a post-hoc explanation tool to a pre-hoc and post-hoc auditing process for enforcing temporal validity in multimodal clinical models.

\section{Methods}

In this work, we use the term auditing to refer to a model development procedure that identifies and suppresses explicit discharge-proxy lexical cues discovered through interpretability analysis. Unlike traditional post-hoc explanation, auditing modifies the training corpus in order to constrain the model's reliance on these cues during retraining. The auditing procedure consists of four steps. First, a base BioClinicalBERT model is fine-tuned on operative-day clinical notes for next-day discharge prediction. Second, token-level attributions are computed using SHAP with prediction probabilities as the explanation target. Third, highly attributed tokens overlapping with a predefined discharge-related lexicon are identified and masked in the training data. Finally, the model is re-trained on the audited corpus using identical optimization settings.

\subsection{Task Definition and Data}

We study next-day discharge prediction following elective spine surgery using retrospective EHR data from a single academic medical center (Fig. 1), comprising 1,956 adult surgical encounters, of which 438 resulted in next-day discharge (LOS=1) and 1,518 had longer hospitalizations (LOS$>$1). The task is formulated as binary classification, where the outcome indicates whether a patient is discharged on the calendar day immediately following surgery (LOS=1) or remains hospitalized longer (LOS$>$1). All inputs were constrained to information available strictly prior to the discharge decision window.

Clinical notes were timestamped and aligned relative to the index surgical date. For patients discharged on postoperative day 1, only notes authored on the day of surgery (DAY 0) were retained. For patients with longer hospitalizations, notes authored on or after the day preceding discharge were excluded. Including earlier notes risks introducing downstream planning information documented retrospectively. No manual lexical filtering or discharge-specific term removal was applied at this stage, allowing the baseline model to capture potentially leakage-prone signals for auditing. The resulting dataset was split at the patient level using stratified sampling.

The study population consists of adult patients undergoing elective spine surgery during routine inpatient care. Clinical documentation includes operative reports, perioperative progress notes, and postoperative assessment notes generated as part of the standard clinical workflow. Note length varies across encounters, ranging from brief procedural summaries to multi-paragraph narratives describing postoperative recovery and care planning. After tokenization using the BioClinicalBERT tokenizer, note sequences were truncated or padded to a maximum input length of 512 tokens for model training.

In addition to textual documentation, the underlying EHR contains structured clinical information such as laboratory measurements, diagnosis codes, procedure codes, medication records, and demographic variables. The present study focuses on clinical notes to examine how language models can learn lexical shortcuts from narrative documentation when predicting near-term hospital discharge. All documentation analyzed in this study was written in English.

\begin{figure}[htbp]
\centering
\includegraphics[width=\columnwidth]{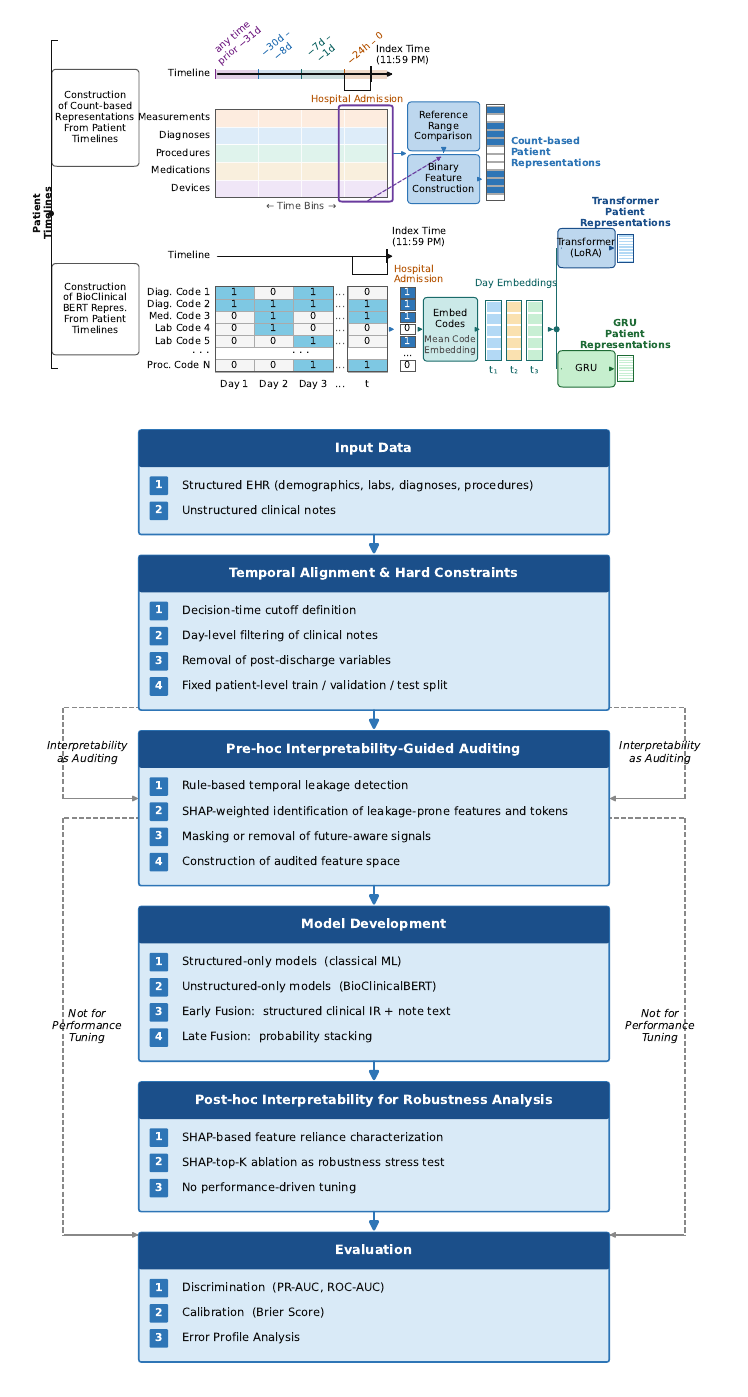}
\caption{Overview of the interpretability-guided auditing framework. (a) Temporal setup for next-day discharge prediction using operative-day clinical notes only. (b) Interpretability-guided auditing pipeline applied to a BioClinicalBERT classifier, where token-level SHAP attribution identifies highly influential discharge-related lexical cues that are masked prior to retraining.}
\label{fig1}
\end{figure}

\subsection{Definition of Auditing}

We define auditing as a three-step process: (1) train an unaudited model under strict temporal filtering to surface candidate dependencies; (2) apply token-level SHAP attribution on training data to identify lexical cues with disproportionately high contribution to positive predictions; (3) suppress these cues via token masking prior to retraining an otherwise identical model. Importantly, the unaudited model is not intended as a deployable predictor, but as an attribution probe to expose potentially invalid dependencies.

\subsection{Interpretability-Guided Pre-hoc Auditing}

We operationalize interpretability as a pre-hoc auditing mechanism rather than a post-hoc explanatory step. Specifically, we apply token-level SHAP attribution to the unaudited baseline model using training data only. SHAP values quantify each token's contribution to the predicted probability of next-day discharge, computed using a fixed, sampling-based approximation with a bounded number of model evaluations. SHAP values were aggregated across training samples to estimate global token importance.

Tokens were flagged for auditing if they satisfied both of the following criteria: (1) high attribution magnitude within the top quantile of mean absolute SHAP values; and (2) lexical alignment with a predefined set of discharge-related terms (e.g., discharge, dc, home, tomorrow). For example, a note containing the phrase ``Likely home tomorrow if pain controlled'' would be transformed during auditing to ``Likely home [MASK] if pain controlled.'' This illustrates how highly attributed discharge-related lexical cues identified through SHAP attribution are masked prior to retraining.

\subsection{Audited Model Retraining and Evaluation}

Following auditing, we retrained the same BioClinicalBERT-LoRA architecture on the audited note corpus, producing the audited model referred to as AuditBERT (Fig. 1). Aside from the masked tokens, all modeling choices, including architecture, optimization settings, and training protocol, were identical to the unaudited baseline. AuditBERT was trained solely on the audited training data and evaluated on held-out validation and test sets that were never used during SHAP computation or auditing.

The auditing procedure is applied only to the training corpus. Validation and test notes remain unchanged and are used solely for downstream evaluation. Consequently, differences between the audited and unaudited models reflect changes in learned model behavior rather than masking of evaluation-time evidence. For evaluation, we report precision-recall behavior, probability calibration (via Brier score and calibration curves), and qualitative differences in attribution patterns between unaudited and audited models. Post-hoc SHAP analysis is used exclusively for diagnostic purposes, to assess whether auditing reduces reliance on explicit discharge-related cues.

\subsection{Unaudited Baseline Model}

As an initial baseline, we fine-tuned BioClinicalBERT on the temporally filtered clinical notes. BioClinicalBERT was chosen due to its domain-specific pretraining on clinical text and prior use in hospital outcome prediction tasks. To ensure computational efficiency and enable repeated retraining under different audit conditions, we applied Low-Rank Adaptation (LoRA), injecting trainable low-rank adapters into the attention layers while freezing the base model parameters. The unaudited model was trained using standard cross-entropy loss and optimized with AdamW. Notes were tokenized using the BioClinicalBERT tokenizer and truncated or padded to a maximum sequence length of 512 tokens.

\section{Results}

Auditing did not consistently improve discrimination metrics such as AUROC or PR-AUC. Instead, the primary effect of the auditing procedure was behavioral. Compared with the unaudited model, the audited model produced more conservative positive predictions and improved probability calibration. These changes suggest reduced reliance on explicit discharge-proxy lexical cues identified through SHAP attribution. Table \ref{tab1} summarizes the performance differences between the unaudited and audited models, highlighting changes in calibration and precision-recall behavior following attribution-guided masking. Performance trade-offs are mainly driven by the minority LOS=1 class, which represents the clinically actionable outcome. In particular, recall for LOS=1 decreased after auditing, reflecting reduced overconfident positives.

\subsection{What Did SHAP Identify in the Unaudited Model?}

The token-level SHAP analysis of the unaudited BioClinicalBERT model revealed a systematic reliance on explicit discharge-related lexical cues. In training samples, tokens such as ``discharge,'' ``dc'' and temporal expressions indicating imminent release consistently appeared among the highest-attribution contributors to positive predictions. This behavior emerged even after strict timestamp-based filtering of clinical notes, indicating that temporal alignment alone is insufficient to eliminate implicit leakage in clinical text. To provide quantitative support beyond illustrative examples, Table \ref{tab1} reports the relative change in calibration error and precision-recall behavior before and after auditing. Although individual token attributions vary between samples, the aggregate effect of suppressing high-attribution discharge proxies is reflected in consistently reduced confidence for near-threshold positive predictions.

\subsection{Effect of Auditing on Model Behavior}

After masking SHAP-identified discharge-related tokens and retraining, AuditBERT exhibited systematic changes aligned with the objectives of interpretability-guided auditing. Post-hoc attribution analysis showed substantial attenuation of explicit discharge terminology, with attribution redistributed to broader contextual descriptions of patient status and care progression. In parallel, predictive behavior shifted to more conservative probability estimates, particularly for borderline cases. This was reflected in improved calibration behavior and more cautious precision-recall trade-offs, with fewer overly confident false-positive discharge predictions. Although raw discrimination metrics occasionally decreased slightly relative to the unaudited model, the audited models demonstrated more stable and interpretable behavior consistent with clinical deployment requirements. Although masking does not guarantee complete removal of all implicit discharge signals, the observed reduction in extreme probability mass and improved calibration curves indicate that auditing systematically weakens dominant leakage pathways rather than merely redistributing them.

\subsection{Structured-Only and Unstructured-Only Ablation Test Results}

Structured-only models achieved strong and stable performance under strict temporal constraints (Table \ref{tab1}). Tree-based methods, particularly Random Forest and LightGBM, demonstrated high ROC-AUC (0.92-0.95), balanced precision-recall behavior, and well-calibrated probability estimates (Brier 0.04-0.06). SHAP analysis showed that procedure codes (CPT) and demographic variables dominated predictions, with limited contribution from laboratory and diagnosis features, suggesting clinically plausible reliance patterns.

In contrast, unstructured-only models showed more fragile behavior. Traditional TF-IDF-based models achieved high accuracy driven by the majority class but suffered from poor recall and low F1 for next-day discharge in several configurations, indicating severe class imbalance sensitivity. Fine-tuned BioClinicalBERT improved contextual modeling but exhibited strong dependence on discharge-related lexical cues prior to auditing, as revealed by SHAP. These results show that unstructured text provides complementary signal but requires explicit auditing to avoid spurious shortcuts.

Although auditing does not uniformly improve ROC-AUC or PR-AUC, audited models consistently exhibit reduced overconfidence, improved calibration, and diminished reliance on explicit discharge-related language. These shifts align with deployment-oriented objectives, where optimistic discrimination driven by temporally invalid cues is undesirable. Performance trade-offs therefore reflect intentional constraint rather than unintended model degradation.

\subsection{Performance Trade-offs and Deployment Validity}

In our results, attribution-guided auditing did not consistently improve conventional discrimination metrics such as ROC-AUC or PR-AUC. In several settings, audited models exhibited modest reductions in these measures relative to unaudited counterparts. Importantly, these changes were accompanied by systematic shifts in model behavior, including reduced overconfidence, improved probability calibration, and decreased reliance on explicit discharge-related lexical cues.

From a deployment perspective, these trade-offs are expected and, in many cases, desirable. Performance gains driven by temporally invalid signals inflate apparent accuracy while undermining generalizability and clinical validity. By suppressing such cues, auditing constrains the hypothesis space available to the model, yielding predictions that are more conservative and better aligned with temporally plausible evidence.

These findings suggest that interpretability-guided auditing should not be evaluated solely by improvements in aggregate discrimination, but by whether model behavior becomes more stable, calibrated, and clinically defensible under realistic documentation conditions. In this sense, observed performance trade-offs reflect intentional constraint rather than degradation and highlight the limitations of accuracy-centric evaluation for high-stakes clinical NLP tasks.

\begin{table*}[t]
\caption{Performance Comparison Under Temporal Auditing}
\begin{center}
\begin{tabular}{lcccccc}
\toprule
Model & AUROC & PR-AUC & Prec. & Rec. & F1 & Brier \\
\midrule
Logistic Regression & 0.91 & 0.63 & 0.58 & 0.54 & 0.56 & 0.061 \\
Random Forest & 0.94 & 0.68 & 0.66 & 0.60 & 0.63 & 0.048 \\
LightGBM & 0.95 & 0.71 & 0.69 & 0.62 & 0.65 & 0.042 \\
TF-IDF + LR & 0.77 & 0.29 & 0.41 & 0.12 & 0.18 & 0.112 \\
BioClinicalBERT (Unaud.) & 0.88 & 0.52 & 0.61 & 0.68 & 0.64 & 0.093 \\
AuditBERT (SHAP-aud.) & 0.85 & 0.49 & 0.66 & 0.57 & 0.61 & 0.071 \\
\bottomrule
\end{tabular}
\label{tab1}
\end{center}
\end{table*}

\section{Discussion}

We note that some discharge-correlated language may reflect legitimate clinical planning rather than spurious artifacts. The auditing procedure does not attempt to perfectly disentangle these cases, but instead targets highly explicit temporal proxies that risk violating temporal validity assumptions. Masking explicit tokens does not guarantee complete removal of all implicit discharge signals, as language models may infer masked content from the surrounding context. Auditing therefore weakens, rather than fully eliminates, leakage pathways. The goal of AuditBERT-Discharge is not to claim complete artifact removal, but to demonstrate that interpretability-guided suppression can measurably alter model behavior toward a more conservative and temporally valid prediction regime.

The lexical cues identified in this study may include both temporally invalid proxies and clinically legitimate planning language. Our auditing framework targets explicit discharge-related expressions that exhibit high attribution in the model but does not fully distinguish between invalid leakage and valid clinical reasoning. Future work should explore more controlled experimental designs that can better separate these mechanisms.

Importantly, AuditBERT-Discharge does not claim to identify causal predictors of discharge, nor to eliminate all forms of spurious correlation. Instead, it serves as a diagnostic and corrective mechanism for highly explicit temporal shortcuts that violate the semantics of the prediction task. In this sense, auditing constrains model behavior rather than optimizing predictive performance, and should be viewed as complementary to, not a replacement for, causal modeling or prospective validation.

Building on this scope clarification, this study reframes interpretability from a post-hoc explanatory tool to a pre-hoc auditing mechanism that constrains what clinical NLP models are permitted to learn. Rather than explaining predictions after training, we use SHAP to identify and suppress leakage-prone signals before final model fitting. This distinction is critical for clinical text, where subtle future-aware language can bypass standard temporal filtering.

AuditBERT-Discharge is intentionally lightweight and practical. By combining token-level SHAP with simple lexical constraints, it enforces temporal validity without large models, external knowledge bases, or architectural changes. More broadly, this work illustrates how interpretability can function as a structural control embedded in the modeling pipeline, emphasizing validity and robustness over optimistic performance.

Due to institutional data restrictions, the clinical dataset used in this study cannot be publicly released. However, the proposed auditing procedure is model- and dataset-agnostic and can be applied to any token-level classifier with access to attribution methods. All methodological details necessary to reproduce the auditing process are provided.

\section{Conclusion}

We introduced AuditBERT-Discharge, a leakage-aware method that operationalizes interpretability as a pre-hoc and post-hoc auditing mechanism for next-day discharge prediction. By integrating SHAP-based attribution with strict temporal constraints, our approach suppresses leakage-prone signals prior to model training. These results highlight the need to embed interpretability into the modeling pipeline to build temporally valid and trustworthy clinical prediction systems.

\section*{Limitations}

This study has several limitations. First, we did not include random masking baselines that would help isolate whether the observed effects are specific to SHAP-identified tokens or reflect general regularization. Second, alternative attribution methods may identify different candidate cues for auditing. Third, replacing tokens with mask symbols may introduce distributional artifacts that could themselves become predictive signals. Future work should examine alternative token removal strategies and controlled experimental designs to better isolate causal mechanisms.

\section*{Ethics Considerations}

All data were de-identified and analyzed under institutional approval. The proposed auditing method aims to reduce misleading model behavior and improve patient safety by discouraging reliance on invalid signals.

\section*{Acknowledgment}

This study was not funded by any external source.

\appendix

\subsection{Additional Experimental Details}

All experiments were implemented in Python using PyTorch, HuggingFace Transformers, PEFT (LoRA), and SHAP, with GPU acceleration when available. Random seeds were fixed across NumPy, PyTorch, and data splitting to ensure reproducibility. For unstructured modeling, BioClinicalBERT served as the base encoder with LoRA applied to attention layers (query and value projections), while base parameters were frozen, reducing trainable parameters to approximately 1 percent and enabling efficient retraining after auditing. The unaudited baseline was trained on temporally filtered clinical notes without lexical exclusions and used as an attribution probe to identify potentially invalid discharge-related dependencies, with token-level SHAP values computed in evaluation mode using only the training data to avoid leakage. Auditing was implemented via token masking, replacing flagged tokens with the model's mask token to preserve sequence structure while suppressing explicit cues, and the audited model reused the same architecture and training configuration with all settings fixed a priori to isolate the effect of attribution-guided masking.

\subsection{Overall Performance Comparison}

Across structured-only, unstructured-only, and audited unstructured settings, models exhibited distinct performance profiles aligned with input modality (Table \ref{tab1}). Structured models showed stable discrimination and calibration under temporal constraints, reflecting the reliability of coded EHR variables. Unstructured models provided complementary context but were more sensitive to temporal leakage. Prior to auditing, models relied on discharge-related lexical cues, inflating apparent performance. After auditing, predictions became more conservative, with attribution redistributed toward clinically plausible context, sometimes with reduced raw discrimination. These findings indicate that interpretability-guided auditing prioritizes temporally valid and clinically meaningful signals over isolated performance gains, supporting the development of reliable clinical NLP systems.

\end{document}